\documentclass[11pt]{article}
\pdfoutput=1
\usepackage[final]{acl}

\usepackage{times}
\usepackage{latexsym}
\usepackage{enumitem}
\usepackage[T1]{fontenc}
\usepackage[utf8]{inputenc}

\usepackage{microtype}

\usepackage{inconsolata}

\usepackage{graphicx}

\usepackage{amsmath}

\usepackage{booktabs}

\title{The Performance--Perception Gap in Hallucination-Risk Visualization}

\author{%
  \textbf{Ryohei\,Kamei\textsuperscript{1}} \quad
  \textbf{Masaki\,Sakata\textsuperscript{1,2}} \quad \\
  \textbf{Asahi\,Hentona\textsuperscript{3,4}} \quad
  \textbf{Kentaro\,Kurihara\textsuperscript{3,4}} \quad
  \textbf{Kentaro\,Inui\textsuperscript{5,1,2}} \quad \\
  $^{1}$ Tohoku University \quad
  $^{2}$ RIKEN \quad
  $^{3}$ CyberAgent \quad
  $^{4}$ AI Shift \quad
  $^{5}$ MBZUAI \\
  \texttt{\{ryohei.kamei.s4, sakata.masaki.s5\}@dc.tohoku.ac.jp}, \\
  \texttt{\{hentona\_asahi, kurihara\_kentaro\}@cyberagent.co.jp} \\
  \texttt{kentaro.inui@mbzuai.ac.ae}
}

\begin{document}

\maketitle

\begin{abstract}
RAG outputs may contain unsupported or source-contradicting claims, requiring user verification.
We examine how the \emph{Information Richness} of hallucination-risk visualizations affects quiz accuracy and perceived usefulness.
In a controlled within-subjects study using pseudo-documents, 12 annotators completed 960 main-experiment trials (2,880 quiz-question responses) across four interfaces that progressively added localization, hallucination-type characterization, and explanations.
Visual cues improved accuracy over no highlighting, but the estimated benefit was largest with type-specific highlighting and did not increase when explanations were added; this pattern does not establish equivalence.
Post-experiment usefulness rankings favored richer interfaces; we describe their divergence from the quiz-accuracy pattern as a descriptive, condition-level performance--perception gap.
An exploratory flipped-label probe provided preliminary evidence that unreliable cues were associated with lower accuracy.
Together, these findings characterize how hallucination-risk visualizations shape performance and perception in a controlled RAG verification task, underscoring the importance of evaluating objective outcomes alongside perceived usefulness.
\end{abstract}

\section{Introduction}
\label{sec:introduction}

Although Retrieval-Augmented Generation (RAG) grounds LLM outputs in external sources and can reduce hallucinations~\cite{Lewis_rag,Fan_ragsurvey_KDD2024}, generated answers may still contain unsupported or source-contradicting claims that mislead users' judgments~\cite{Augenstein2024,yin-etal-2023-large,Spatharioti_CHI2025,niu-etal-2024-ragtruth,shi-etal-2024-trusting}.
To address this problem, prior work has studied model- and system-level approaches for detecting and mitigating hallucinations~\cite{huang_hallucination_survey, ji_hallucination_survey_ACM2023, rawte-etal-2023-troubling, manakul-etal-2023-selfcheckgpt, zhang-etal-2023-enhancing-uncertainty}, as well as interface-level cues that help users assess output reliability~\cite{Spatharioti_CHI2025,Cheng_RELIC_CHI2024}.
For example, systems such as SummVis~\cite{summvis} and HILL~\cite{hill} support user verification by highlighting spans whose factuality cannot be guaranteed.

\begin{figure}[t]
\centering
\includegraphics[width=\columnwidth]{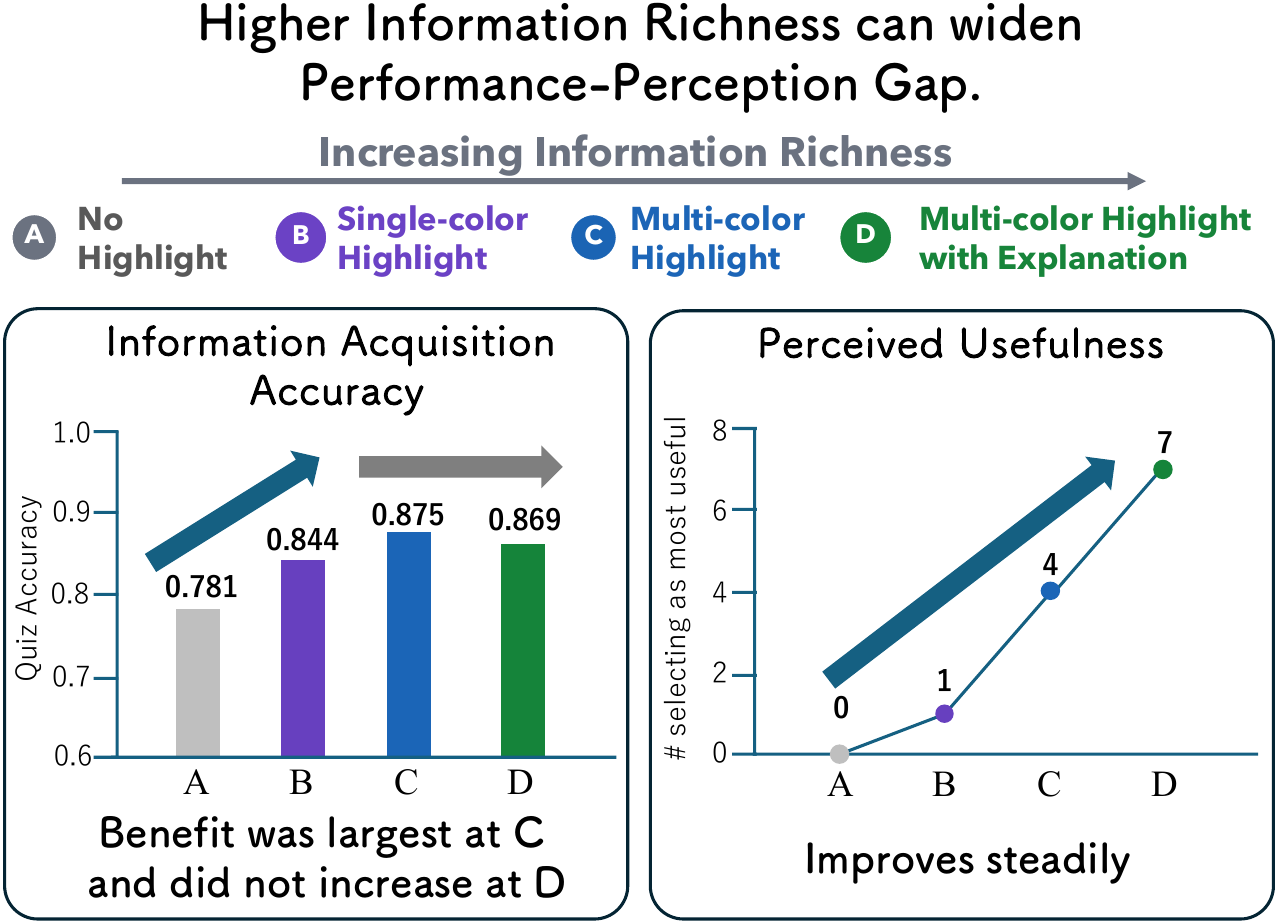}
\caption{
Overview of this paper.
Post-experiment usefulness rankings favored richer interfaces, whereas the estimated quiz-accuracy benefit was largest for localization plus characterization (C) and did not increase after explanations were added (D).
}
\label{fig:summary}
\end{figure}

\begin{figure*}[t]
\centering
\includegraphics[width=\linewidth]{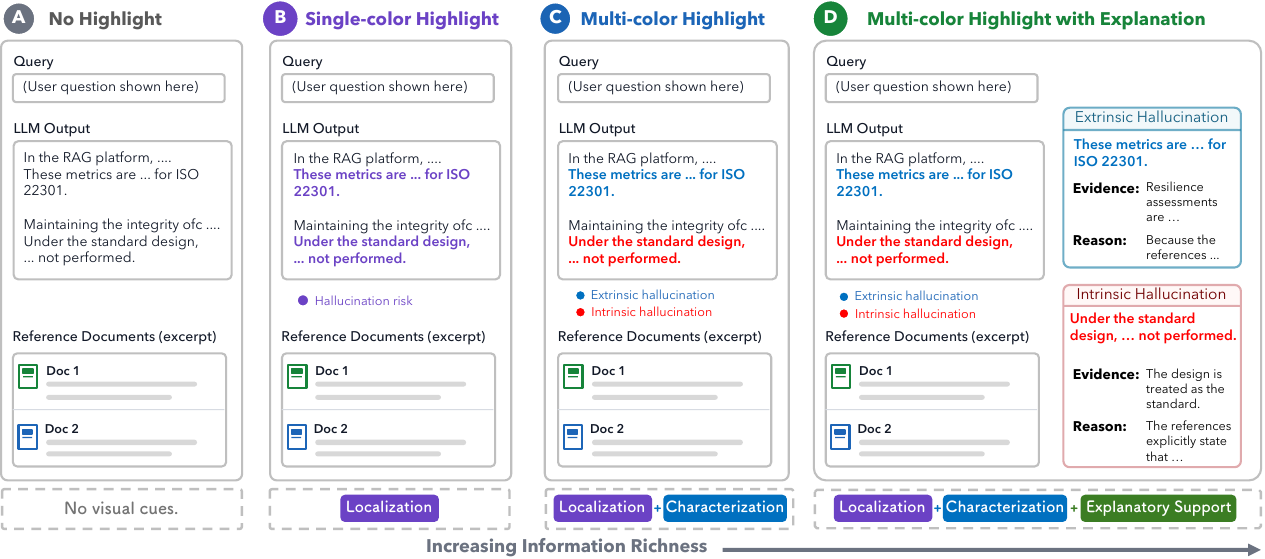}
\caption{
Illustration of the four visualization methods: A: \textit{No Highlight}, B: \textit{Single-color Highlight}, C: \textit{Multi-color Highlight}, D: \textit{Multi-color Highlight with Explanation}.
The experiment was conducted in Japanese, but this figure shows English examples.
}
\label{fig:methods_vis}
\end{figure*}

Despite these advances, existing research has not sufficiently examined what information hallucination visualization should present or in how much detail~\cite{Reinhard_fact_or_fiction}.
With too little information, users may overlook which parts require verification.
Richer cues, such as fine-grained color coding or textual reasons for hallucination judgments, may make an interface appear more transparent and actionable, thereby increasing perceived usefulness.
However, they may also increase cognitive cost or encourage users to substitute interface explanations for independent source inspection~\cite{dilemma}.
Thus, an important open question is how the amount and specificity of verification-relevant information affect users' actual information acquisition and subjective evaluations.

To address this gap, we focus on \emph{Information Richness} in hallucination-risk visualization, defined as the amount and specificity of verification-relevant information that an interface presents about an LLM output.
Specifically, we ask how Information Richness affects \textbf{(RQ1)} users' quiz-based information acquisition accuracy and \textbf{(RQ2)} their subjective evaluations.
We investigate these RQs in a controlled, quiz-based RAG verification experiment using pseudo-documents and four visualization methods (\textit{No Highlight}, \textit{Single-color Highlight}, \textit{Multi-color Highlight}, and \textit{Multi-color Highlight with Explanation}) that present progressively richer information to annotators (Figure~\ref{fig:methods_vis}).

Our results show that visual cues improved quiz accuracy relative to \textit{No Highlight}. However, the estimated benefit was largest for \textit{Multi-color Highlight}, which additionally characterized hallucination types, and did not increase further with explanatory support.
In contrast, post-experiment usefulness rankings favored richer interfaces.
We describe the differing condition-level patterns in relative usefulness preferences and quiz-accuracy estimates as a descriptive performance--perception gap (Figure~\ref{fig:summary}).
In a separate exploratory robustness probe conducted after the main experiment, intentionally flipped hallucination labels were associated with lower quiz accuracy.
This preliminary result highlights hallucination detector reliability as an important boundary condition and motivates conservative presentation when hallucination judgments are uncertain.
Overall, this study provides evidence about differences among the tested interfaces within a controlled, quiz-based RAG verification task.
The 12-annotator, within-subjects design supports these within-study comparisons; naturalistic use and population-level variation in preferences, trust formation, and behavioral mechanisms remain beyond the scope of the present evidence.
\footnote{The dataset we constructed (consisting of the pseudo-document corpus and the quiz-style tasks) will be curated and released in the near future.}

\section{Related Work}
\label{sec:related_work}

\subsection{Hallucinations and Verifiability in RAG}
Prior work has proposed taxonomies for unsupported or incorrect LLM generation~\cite{huang_hallucination_survey,ji_hallucination_survey_ACM2023,Augenstein2024}.
A common distinction is between intrinsic hallucinations (IH), which contradict available evidence, and extrinsic hallucinations (EH), which are unsupported by the provided context~\cite{huang_hallucination_survey}.
In RAG settings, IH can be checked against retrieved evidence, whereas EH signals that the context is insufficient, motivating our use of this distinction as verification-relevant characterization.

\subsection{Hallucination Visualization Interfaces}
Prior interfaces support verification by highlighting questionable spans, exposing evidence, presenting factuality scores or source attributions, and visualizing response stability~\cite{hill,do2024facilitating_chiws,summvis,Cheng_RELIC_CHI2024}; related work also considers LLM assistance for long documents~\cite{ni-etal-2023-chatreport}.
Most studies compare complete interfaces with baselines rather than ablating individual presentation elements, and often treat hallucinations as a single category.
Consequently, the effects of distinguishing IH from EH and of presenting incorrect cues remain unclear.

\subsection{Explanation and Overreliance}
Explanations and evidence can support decisions but may also increase time and cognitive load or induce excessive confidence.
RAG studies report that explanations can improve hallucination detection with added overhead~\cite{Reinhard_fact_or_fiction}, while other work finds that perceived helpfulness need not yield performance gains and may help justify incorrect answers~\cite{dilemma}.
LLM-based search and cues such as explanations, citations, and inconsistency signals also shape decision speed, accuracy, and reliance~\cite{Spatharioti_CHI2025,Kim_overreliance_CHI2025}.
Yet controlled comparisons of lightweight cues remain limited, particularly regarding how their trade-offs change when combined with explanations.

\section{Definition of Information Richness in Hallucination-Risk Visualization}
\label{sec:information_richness}

We define \emph{Information Richness} as the amount and specificity of verification-relevant information that an interface presents about an LLM output: where a potential problem appears, what kind of problem is suspected, and how to interpret the cue.
Drawing on prior visualization and factuality work, we operationalize it through three recurring aspects: \emph{localization}, which marks potentially problematic content~\cite{summvis,hill}; \emph{characterization}, which identifies the suspected issue through labels or related reliability indicators~\cite{hill,niu-etal-2024-ragtruth}; and \emph{explanatory support}, which helps users interpret and act on a cue~\cite{hill,Reinhard_fact_or_fiction}.
These aspects form a practical, rather than exhaustive, synthesis for controlled comparison.

Figure~\ref{fig:methods_vis} implements them cumulatively: \textit{No Highlight} is the baseline; \textit{Single-color Highlight} adds localization; \textit{Multi-color Highlight} adds IH/EH characterization~\cite{huang_hallucination_survey}; and \textit{Multi-color Highlight with Explanation} adds brief explanations.
Thus, Information Richness increases in the order \textit{No Highlight} $<$ \textit{Single-color Highlight} $<$ \textit{Multi-color Highlight} $<$ \textit{Multi-color Highlight with Explanation}, while the underlying detector outputs remain fixed.

\section{Experimental Setup}
\label{sec:Experiment}
We conducted a controlled quiz-based experiment to examine how stepwise increases in Information Richness (\S~\ref{sec:information_richness}) affect annotators' quiz-based information acquisition, subjective evaluations, and time cost.
Each of 100 samples comprised a query, an LLM response, and three multiple-choice comprehension questions whose answers were uniquely determined by the reference documents.
Annotators could consult the references and completed each sample under a five-minute limit; we recorded quiz accuracy and response time and administered questionnaires after each method and after the experiment.
Following prior reading-comprehension evaluations~\cite{agrawal-carpuat-2024-meaning,laban-etal-2021-keep}, we treat quiz accuracy as task-specific information acquisition rather than a comprehensive measure of naturalistic verification behavior.
The predefined queries, time limit, and single-turn design support controlled comparison but do not reproduce all aspects of naturalistic RAG use.

\subsection{Data Preparation}
\label{subsec:data_preparation}
This subsection describes how we constructed the 100-sample problem set for annotation.
First, we built a pseudo-document corpus using an LLM for a RAG setting, consisting of 100 pseudo-document samples.
The documents describe fictional internal-company information so that annotators would be unlikely to answer from prior knowledge and would instead need to use the presented RAG responses and reference documents.
The corpus covers 20 main topics, including organizational structure, information security, financial management, and AI ethics (see Appendix~\ref{sec:appendix_corpus_contents} for the full list of main topics).
Next, using this corpus as input, we generated queries about the document contents and corresponding LLM responses.
To include situations that require verification, we intentionally inserted up to one IH and up to one EH per generated response, yielding samples with none, one (IH or EH), or both (one each).
The 100 samples comprised 37 with neither IH nor EH, 18 with one IH only, 24 with one EH only, and 21 with both one IH and one EH each.
We then created three multiple-choice comprehension questions for each query--response pair, using the corpus above as the information source (see Appendix~\ref{sec:appendix_annotation_ui}).
Thus, the task set systematically covers the four hallucination configurations relevant to our RQs, but its 100 samples do not span the full diversity of real-world topics, document formats and lengths, or evidential ambiguity.

We fed the resulting LLM responses into a hallucination detection model, which labeled each sentence as ``supported by the source documents,'' ``contradicts the source (IH),'' or ``cannot be verified from the source (EH).''
The distribution of the prediction results is shown in Appendix~\ref{sec:appendix_hallucination_distribution}.
We constructed the visualization methods, \textit{Single-color Highlight}, \textit{Multi-color Highlight}, and \textit{Multi-color Highlight with Explanation}, so that they used the same predicted labels and varied only the presentation style.
Against the 605 sentence-level labels used during data construction, the GPT-5 detector achieved accuracy $0.9653$ and macro-F1 $0.9304$ (Table~\ref{tab:detector_performance}).
Detector errors affect the cues shown in Methods B--D; because all three methods use exactly the same predictions, however, their differences cannot be attributed to different detector outputs.
In this study, we used GPT-5~\cite{openai-gpt5-system-card} as both the text generation model and the hallucination detection model.

\begin{table}[t]
\centering
\small
\setlength{\tabcolsep}{4pt}
\begin{tabular}{lrrr}
\toprule
\multicolumn{4}{l}{\textit{(a) Confusion matrix (counts)}} \\
Gold $\backslash$ Predicted & Supported & IH & EH \\
\hline
Supported & 482 & 2 & 17 \\
IH & 0 & 48 & 0 \\
EH & 0 & 2 & 54 \\
\toprule
\multicolumn{4}{l}{\textit{(b) Per-class performance}} \\
Class & Precision & Recall & F1 \\
\hline
Supported & 1.0000 & 0.9621 & 0.9807 \\
IH & 0.9231 & 1.0000 & 0.9600 \\
EH & 0.7606 & 0.9643 & 0.8504 \\
\bottomrule
\end{tabular}
\caption{GPT-5 sentence-level hallucination-detector performance against the labels used during data construction ($N{=}605$ sentences; accuracy: $0.9653$; macro-F1: $0.9304$). Rows in the confusion matrix are gold labels and columns are detector predictions.}
\label{tab:detector_performance}
\end{table}

\subsection{User Experiment Procedure}
\label{subsec:annotation_procedure}

We treated annotators' task of answering three quiz questions per sample while viewing the RAG system output as ``annotation,'' and collected quiz correctness, response time, interaction logs, and subjective evaluations.
For each sample, annotators viewed the query, LLM response, three multiple-choice quiz questions, time limit, and reference documents accessible via a reference button.
We also recorded whether annotators clicked the reference button.
The annotation interface is shown in Appendix~\ref{sec:appendix_annotation_ui}.
For all visualization methods, including the exploratory robustness probe, annotators were instructed that highlighted spans were not necessarily errors but reference cues, and were encouraged to consult the reference documents when needed.

Subjective evaluations consisted of two questionnaires.
The during-experiment questionnaire was designed based on NASA-TLX, a measure of subjective workload~\cite{hancock_development_1988}, and prior hallucination-visualization UI studies~\cite{do2024facilitating_chiws,hill}.
The post-experiment questionnaire was administered after annotators experienced all methods and included ranking of each visualization method's usefulness.
Questionnaire items are listed in Appendix~\ref{sec:questionnaire_question}.
As supplementary descriptive evidence about the task set, all 12 annotators indicated that the problems were difficult to some degree.
Eight agreed that the problems were approximately equal in difficulty, whereas four disagreed, suggesting reasonable but imperfect consistency.

We compared four visualization methods (Figure~\ref{fig:methods_vis}).
\textit{No Highlight} showed only the LLM response without hallucination-related cues.
\textit{Single-color Highlight}, based on the hallucination detection model outputs, highlighted sentences suspected of hallucination in a single color, without distinguishing IH from EH.
\textit{Multi-color Highlight} used different colors for IH and EH.
\textit{Multi-color Highlight with Explanation} further added brief textual explanations below the response, describing why each sentence was judged as IH or EH.
Thus, the methods progressively increased Information Richness presented to annotators.

We recruited 12 annotators for a controlled within-subjects experiment.
Each annotator experienced all four visualization methods, enabling direct comparison of Information Richness while reducing between-annotator variance.
Each annotator completed 20 samples per method, yielding 80 sample-level trials per annotator and 960 trials in the main experiment.
Because each sample contained three quiz questions, the main experiment produced 2,880 quiz-question responses.
Including the exploratory robustness probe, the full dataset comprises 3,600 quiz-question responses.

To improve the validity of condition comparisons, we used a Latin-square assignment over visualization methods and sample subsets to mitigate the effects of order, learning, and fatigue.
Specifically, for annotators $P_1$--$P_{12}$, visualization methods $A$: \textit{No Highlight}, $B$: \textit{Single-color Highlight}, $C$: \textit{Multi-color Highlight}, $D$: \textit{Multi-color Highlight with Explanation}, and sample subsets $S_1$--$S_4$, we assigned conditions as follows:

\vspace{-2mm}
\[
\begin{aligned}
&\text{\textbullet\ } P_1\text{: }(\text{A, }S_1)\rightarrow(\text{B, }S_2)\rightarrow(\text{D, }S_3)\rightarrow(\text{C, }S_4)\\
&\text{\textbullet\ } P_2\text{: }(\text{B, }S_1)\rightarrow(\text{C, }S_2)\rightarrow(\text{A, }S_3)\rightarrow(\text{D, }S_4)\\
&\text{\textbullet\ } P_3\text{: }(\text{C, }S_1)\rightarrow(\text{D, }S_2)\rightarrow(\text{B, }S_3)\rightarrow(\text{A, }S_4)\\
&\text{\textbullet\ } P_4\text{: }(\text{D, }S_1)\rightarrow(\text{A, }S_2)\rightarrow(\text{C, }S_3)\rightarrow(\text{B, }S_4)\\
&\text{\textbullet\ } \text{(} P_5 \text{--} P_{12} \text{ repeat the above)}
\end{aligned}
\]
This assignment cannot fully separate residual fatigue from subset-related variation within a single block.

For the main experiment, models of quiz accuracy, reference-view counts, timeout risk, and completion time were fitted at the annotator--sample trial level (960 trials; completion time used the 855 non-timeout trials).
Quiz accuracy was analyzed with a grouped-binomial generalized linear mixed model (GLMM) on the number of correct responses out of the three quiz questions in each trial, reference-view counts with a negative binomial GLMM, timeout risk with a logistic GLMM, and completion time with a linear mixed model on log time.
Each model included visualization method as a categorical fixed effect, presentation order as a numeric covariate, and random intercepts for annotator and sample (the reference-view model retained only the annotator intercept).
Overall method effects were tested with likelihood-ratio tests, and Holm correction was applied within each family of follow-up comparisons reported with the corresponding results.
Full specifications, estimation procedures, and diagnostics are given in Appendix~\ref{sec:appendix_model_details}.
The study was not preregistered.
The RQs, outcome measures, and intention to compare quiz accuracy across methods were specified before analysis; the human-performance-ceiling and quiz-question-allocation checks, behavioral-mechanism interpretation, and Method E are explicitly exploratory.

\section{Results and Discussion}
\label{sec:results}
We organize the results around the two research questions and then synthesize them as a gap between perceived usefulness and quiz-based information-acquisition performance.
RQ1 examines whether increasing Information Richness through localization, characterization, and explanatory support improves quiz accuracy (\S~\ref{subsec:RQ1}).
For RQ2, we examine post-experiment usefulness rankings, with the during-experiment questionnaire as additional descriptive evidence (\S~\ref{subsec:RQ2}).
We then discuss reference-view counts and task time as behavioral and cost measures behind this gap (\S~\ref{subsec:gap}).
Finally, we report a separately conducted exploratory robustness probe that examines what may happen when the visualization labels themselves are unreliable (\S~\ref{subsec:additional_experiment}).

\begin{table*}[t]
\centering
\setlength{\tabcolsep}{4pt}
\begin{tabular}{lcccc@{\hspace{1em}}ccc}
\hline
 & & \multicolumn{3}{c}{Compared with A} & \multicolumn{3}{c}{Compared with previous method} \\
\cline{3-5}\cline{6-8}
Method & Accuracy & $\Delta$ & OR [95\% CI] & $p_{\text{Holm}}$ & $\Delta$ & OR [95\% CI] & $p_{\text{Holm}}$ \\
\hline
A & 0.781 & -- & 1 (ref) & -- & -- & -- & -- \\
B & 0.844 & $+0.063$ & 1.68 [1.16, 2.42] & $<0.001$ & $+0.063$ & 1.68 [1.16, 2.42] & $0.002$ \\
C & 0.875 & $+0.094$ & 2.25 [1.52, 3.32] & $<0.001$ & $+0.031$ & 1.34 [0.89, 2.01] & $0.17$ \\
D & 0.869 & $+0.088$ & 2.17 [1.48, 3.19] & $<0.001$ & $-0.006$ & 0.97 [0.63, 1.47] & $0.84$ \\
\hline
\end{tabular}
\caption{
Quiz accuracy by visualization method.
Accuracy is the annotator-averaged descriptive mean; $\Delta$ denotes descriptive differences.
OR (odds ratio) estimates and $p$-values are from a grouped-binomial GLMM.
Holm correction was applied separately to the three comparisons with Method A (B/A, C/A, D/A) and the three adjacent comparisons (B/A, C/B, D/C), and 95\% confidence intervals (CIs) are Bonferroni-adjusted within each of these families.
Previous-method comparisons use the immediately less rich method.
}
\label{tab:RQ1_results}
\end{table*}

\subsection{RQ1: Does Information Richness improve quiz-based information acquisition accuracy?}
\label{subsec:RQ1}

Because hallucination-risk visualization is intended to support accurate source-grounded information acquisition, RQ1 focuses on task-specific quiz accuracy.
Quantitative results for the quiz task are shown in Table~\ref{tab:RQ1_results}.
Table~\ref{tab:RQ1_results} reports descriptive accuracy, comparisons with Method A (\textit{No Highlight}), and adjacent contrasts between successive levels of Information Richness.
The trial-level grouped-binomial GLMM provided strong evidence for an overall method effect on quiz accuracy ($p<0.001$).
Compared with \textit{No Highlight}, each of the three richer methods improved accuracy with strong evidence (each $p_{\text{Holm}}<0.001$): mean accuracy rose from $0.781$ under \textit{No Highlight} to $0.844$ under \textit{Single-color Highlight} ($\Delta=+0.063$), $0.875$ under \textit{Multi-color Highlight} ($\Delta=+0.094$), and $0.869$ under \textit{Multi-color Highlight with Explanation} ($\Delta=+0.088$).
However, this improvement over \textit{No Highlight} did not grow monotonically with Information Richness: the model-based effect size, measured as the odds ratio (OR) relative to \textit{No Highlight} in Table~\ref{tab:RQ1_results}, peaked at \textit{Multi-color Highlight} and was not larger at \textit{Multi-color Highlight with Explanation}, with the descriptive gain following the same pattern (largest at \textit{Multi-color Highlight}, $\Delta=+0.094$; slightly smaller at \textit{Multi-color Highlight with Explanation}, $\Delta=+0.088$).
Adjacent contrasts provided clear evidence only for \textit{Single-color Highlight} vs.\ \textit{No Highlight} ($p_{\text{Holm}}=0.002$), with no clear evidence for either the \textit{Multi-color Highlight} vs.\ \textit{Single-color Highlight} contrast ($p_{\text{Holm}}=0.17$) or the \textit{Multi-color Highlight with Explanation} vs.\ \textit{Multi-color Highlight} contrast ($p_{\text{Holm}}=0.84$).
The C--D contrast does not establish equivalence or exclude a meaningful additional benefit of explanations; it shows only that no additional measured benefit was observed in this study.
Thus, with respect to RQ1, the data suggest that all three cue-based methods improve quiz accuracy over no cue, while the estimated benefit was largest at C and did not increase further at D.

\begin{figure}[t]
\centering
\includegraphics[width=\columnwidth]{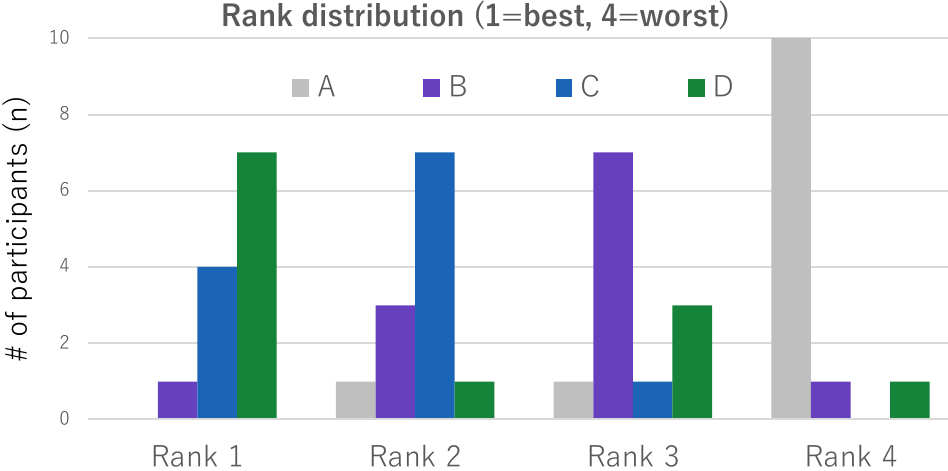}
\caption{
Post-experiment usefulness rankings of Methods A--D ($n=12$).
}
\label{fig:methods_usefulness_ranking}
\end{figure}

\begin{figure*}[t]
\centering
\includegraphics[width=\linewidth]{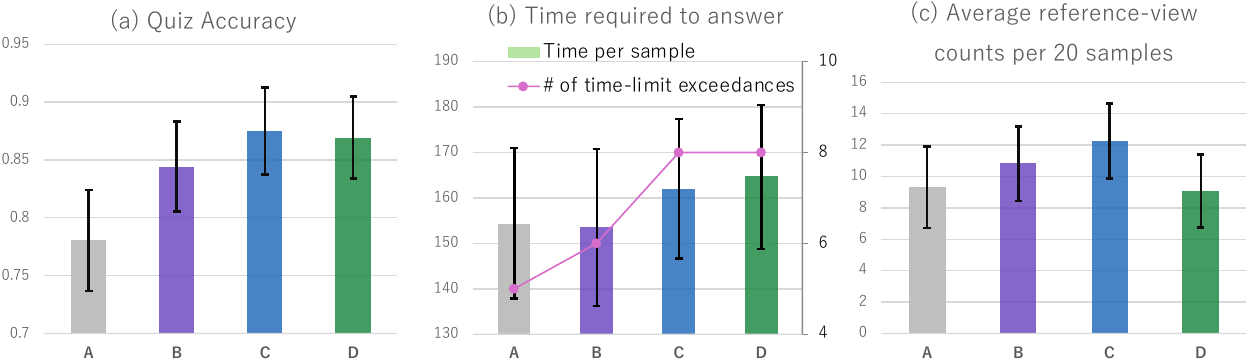}
\caption{
Descriptive summary of main-experiment outcomes.
We computed within-method averages for each annotator and report means across annotators.
(a) Quiz accuracy.
(b) Time per sample for non-timeout samples (bars) and time-limit exceedances (line; right axis).
(c) Average reference-view counts per 20 samples as a process measure of verification behavior.
Error bars show variability across annotators.
}
\label{fig:results}
\end{figure*}

\subsection{RQ2: Does Information Richness improve subjective evaluations?}
\label{subsec:RQ2}

\begin{table}[t]
\centering
\small
\setlength{\tabcolsep}{4pt}
\resizebox{\columnwidth}{!}{%
\begin{tabular}{lcccc}
\hline
 & \multicolumn{4}{c}{Mean (SD)} \\
\cline{2-5}
Item & A & B & C & D \\
\hline
Q1 & $4.33$ ($1.87$) & $5.00$ ($1.35$) & $4.67$ ($1.23$) & $5.08$ ($1.38$) \\
Q2 & $3.33$ ($1.67$) & $3.00$ ($1.60$) & $3.33$ ($1.56$) & $3.58$ ($1.62$) \\
Q3 & $4.42$ ($1.24$) & $3.67$ ($1.72$) & $4.17$ ($1.34$) & $4.33$ ($1.44$) \\
Q4 & $4.58$ ($2.15$) & $4.83$ ($1.64$) & $4.33$ ($1.67$) & $4.58$ ($1.78$) \\
Q5 & $5.25$ ($1.36$) & $5.42$ ($1.73$) & $5.92$ ($0.67$) & $5.42$ ($1.00$) \\
Q6 & $5.00$ ($1.21$) & $5.00$ ($1.41$) & $4.75$ ($1.29$) & $4.83$ ($1.47$) \\
\hline
\end{tabular}
}
\caption{
Descriptive statistics for during-experiment questionnaire ratings by visualization method ($n=12$).
}
\label{tab:RQ2_questionnaire}
\end{table}

Beyond objective information acquisition, hallucination-risk visualization may also shape annotators' subjective experience, including how useful, trustworthy, and burdensome the system feels.
RQ2 therefore examines whether increasing Information Richness leads to more favorable subjective evaluations and task experience.
Figure~\ref{fig:methods_usefulness_ranking} summarizes the post-experiment ranking results in which annotators comparatively ranked the four methods by perceived usefulness.
These rankings showed a clear preference for information-rich methods.
A Friedman test provided strong evidence for an overall method effect ($p<0.001$), and an exact Page trend test provided strong evidence for an ordered pattern corresponding to increasing Information Richness, \textit{No Highlight} $<$ \textit{Single-color Highlight} $<$ \textit{Multi-color Highlight} $<$ \textit{Multi-color Highlight with Explanation} ($p<0.001$).
For RQ2, annotators ranked interfaces with greater Information Richness as more useful.

The during-experiment questionnaire responses (Table~\ref{tab:RQ2_questionnaire}) did not provide clear evidence for method effects, so we interpret the ratings descriptively.
Trust was higher under \textit{Multi-color Highlight with Explanation} and \textit{Single-color Highlight}; workload was lowest under \textit{Single-color Highlight} and highest under \textit{Multi-color Highlight with Explanation}; ease of judging correctness showed no consistent ordering; and critical evaluation peaked under \textit{Multi-color Highlight}.
Thus, these items did not add clear evidence for RQ2 beyond the rankings, but suggest potentially different roles for the components of Information Richness: characterization may promote critical engagement, while explanatory support may increase trust, usefulness, and workload.

\subsection{Interpreting the Performance--Perception Gap}
\label{subsec:gap}

\paragraph{Performance--perception gap and exploratory checks.}
Taken together, \S~\ref{subsec:RQ1} and \S~\ref{subsec:RQ2} indicate different condition-level patterns: post-experiment usefulness rankings favored richer methods, whereas the estimated quiz-accuracy benefit over \textit{No Highlight} was largest at \textit{Multi-color Highlight} and did not increase at \textit{Multi-color Highlight with Explanation} (Figure~\ref{fig:results}(a); Table~\ref{tab:RQ1_results}).
We use \emph{performance--perception gap} descriptively to refer to this discrepancy between relative usefulness preferences and quiz-accuracy estimates.
We did not directly test the within-annotator relationship between perceived usefulness and quiz accuracy.
The forced rankings do not capture ties, preference magnitudes, or absolute usefulness.
This comparison does not establish an accuracy plateau or equivalence between C and D.

Prompted by the observed pattern, we explored two accounts based on quiz-question properties; these checks were not preregistered.
To explore a possible human-performance ceiling account, an exploratory quiz-question-level model provided strong evidence that quiz questions with lower accuracy under Methods A--C were more likely to elicit errors under Method D ($p<0.001$; Appendix~\ref{sec:appendix_exploratory_models}).
However, these errors were not confined to hard quiz questions: among the 70 of 240 quiz questions with at least one incorrect response under Method D, 49 (70.0\%) were still answered correctly by more than half of the annotators under Method D, and 52 (74.3\%) under Methods A--C.
For a quiz-question allocation account, a separate exploratory quiz-question-level model showed no clear evidence for a C--D difference ($p_{\text{Holm}}=0.803$).
These exploratory checks did not identify quiz-question difficulty or allocation as sufficient to explain the observed C--D pattern, but do not establish equivalence.

\paragraph{Exploratory behavioral interpretation.}
Reference-view counts, as a process measure of verification behavior, descriptively peaked at \textit{Multi-color Highlight} ($12.25$ per 20 samples) and dropped at \textit{Multi-color Highlight with Explanation} ($9.08$; Figure~\ref{fig:results}(c)).
A sample-level model provided weak evidence for an overall method effect ($p=0.06$), and this evidence depended on the count-distribution assumption (Appendix~\ref{sec:appendix_estimation_checks}).
The questionnaire item on critical evaluation (Q5; Table~\ref{tab:RQ2_questionnaire}) showed a descriptively similar pattern, with the highest score under \textit{Multi-color Highlight}.
Reference-button clicks indicate only whether a document was opened; they do not show whether it was read carefully, understood, or used in the final answer.
Several free-text responses nevertheless provide qualitative context consistent with reduced inspection effort: annotators reported that explanations reduced effort, enabled immediate answers, or could serve as excerpts of relevant evidence.
One possible interpretation is therefore that characterization made cues more actionable, whereas explanatory support sometimes compressed or partially substituted for independent source inspection.
This mechanism remains tentative and would require richer process measures, such as the passages viewed, within-document scrolling, dwell time, or eye tracking, to test directly.
Task time, as a cost measure, provided complementary evidence (Figure~\ref{fig:results}(b)): a two-part analysis did not provide clear evidence that timeout risk differed across methods but did provide strong evidence that completion time among non-timeout sample-level trials differed across methods.
The clearest contrast was between \textit{Single-color Highlight} and \textit{Multi-color Highlight}, where adding characterization required about 11\% more time on completed trials (Holm-adjusted $p=0.005$), consistent with a cost of interpreting different cue types; adding explanatory support did not clearly increase completion time beyond characterization, and no additional measured accuracy benefit was observed.

\paragraph{Interpretation and design implication.}
This gap is consistent with prior work on the explanation dilemma~\cite{dilemma}, which argues that explanations for incorrect AI predictions can strengthen feelings of usefulness and reassurance without necessarily promoting verification behavior, and may instead induce overreliance and omitted verification.
It is also compatible with resource-rational accounts, in which people allocate limited time, attention, and cognitive effort by trading off accuracy or utility against cognitive cost~\citep{lieder2020}: localization may reduce the search cost of finding where verification is needed, characterization may increase the information scent~\citep{pirolli1999} and actionability of cues for selective source checking, and explanatory support may act as an external scaffold, consistent with cognitive-offloading work~\citep{sachdeva2020}.
These theories offer possible interpretations rather than causal explanations established by our data.
Within this controlled task, \textit{Multi-color Highlight}, which combines localization and characterization without explanatory support, was a promising design candidate: it achieved the highest descriptive accuracy, elicited the most reference views descriptively, and showed the highest descriptive score for critical evaluation, whereas post-experiment usefulness rankings favored richer methods even though the estimated quiz-accuracy benefit did not increase at D.
This pattern should not be read as a general recommendation against explanations, because their effects may differ with other tasks, interfaces, detectors, and user populations.

\begin{figure}[t]
\centering
\includegraphics[width=\linewidth]{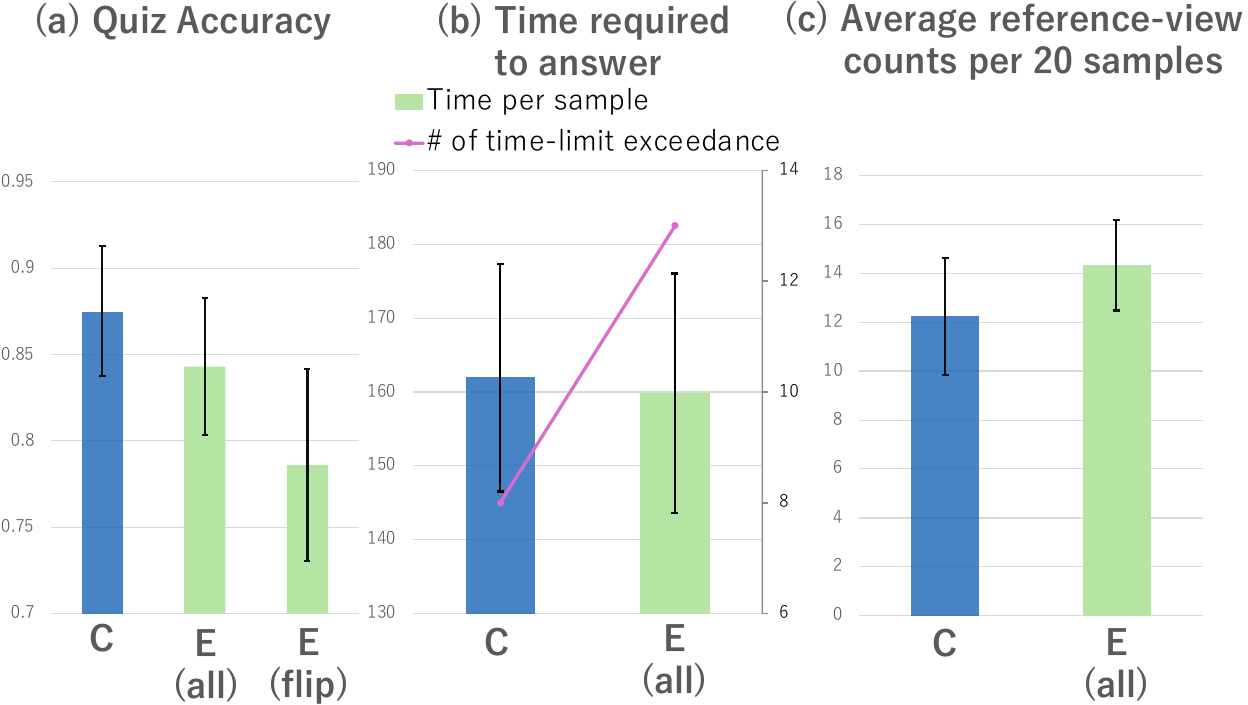}
\caption{
Exploratory robustness-probe results for Method C (defined in Figure~\ref{fig:methods_vis}) and Method E (\textit{Multi-color Highlight with Flipped Labels}).
Values are annotator-level within-method averages, averaged across annotators.
}
\label{fig:ex2_results}
\end{figure}

\subsection{Exploratory Robustness Probe with Incorrect Visualization}
\label{subsec:additional_experiment}

In practice, hallucination detection models may make incorrect judgments, and prior work has reported that adding textual explanations can induce overreliance~\cite{dilemma}.
However, the extent to which highlight-based visualization with incorrect labels may mislead annotators has not been extensively examined.
To explore this possibility, we conducted a robustness probe outside the Latin-square design of Methods A--D, introducing Method E: \textit{Multi-color Highlight with Flipped Labels}.
It used the same Information Richness and UI as \textit{Multi-color Highlight} but intentionally flipped hallucination-judgment labels for 10 of the 20 samples.
Method E was administered after all annotators had completed Methods A--D, on a separate 20-sample subset $S_5$ (Appendix~\ref{sec:appendix_hallucination_distribution}) that did not overlap with $S_1$--$S_4$ used in the main experiment.
As in the main experiment, annotators were instructed that visual cues are imperfect.
Because Method E (all) thus mixes ``correct (non-flipped)'' and ``incorrect (flipped)'' visualizations (10 samples each), we analyzed the data from two complementary perspectives: (i) an annotator-level comparison between Methods C and E with non-identical sample sets, and (ii) a within-E comparison between flipped- and non-flipped-label samples, holding the interface and annotator pool fixed but not the items.

In the annotator-level comparison (Wilcoxon signed-rank tests), we did not obtain clear evidence of a difference in quiz accuracy ($p=0.23$; descriptive: $0.875$ for \textit{Multi-color Highlight} vs.\ $0.843$ for Method E (all); Figure~\ref{fig:ex2_results}(a)), but the data provided suggestive but inconclusive evidence that Method E elicited more reference views ($p=0.0987$; Figure~\ref{fig:ex2_results}(c)).
Within Method E, a sample-level grouped-binomial GLMM showed lower quiz accuracy for flipped-label than non-flipped-label samples (likelihood-ratio test, $p=0.04$; descriptive: $0.786$ vs.\ $0.900$), with no clear evidence of effects on reference-view counts, timeout risk, or non-timeout completion time (Appendix~\ref{sec:appendix_method_e_models}).
Because label status was not randomized over matched items, Method E was always administered after the main conditions, and the C--E comparison used different subsets, item, fatigue, learning, and carry-over effects cannot be separated fully from label flipping.
Given that hallucination detection remains challenging (F1 score of only 0.7109~\cite{sadat-etal-2023-delucionqa}), this exploratory result provides preliminary evidence that visualization may hinder quiz-based information acquisition when the underlying labels are unreliable; the main performance--perception finding does not depend on this probe.

\section{Conclusion}
\label{sec:conclusion}

This study examined how Information Richness of hallucination-risk visualization in a RAG system affects annotators' quiz-based information acquisition, verification behavior, time cost, and subjective evaluations in a controlled verification task.
Across four methods, hallucination cues improved quiz accuracy over the no-highlight baseline, but the estimated benefit was largest for type-specific highlighting and did not increase with explanatory support.
This C--D contrast neither establishes equivalence nor excludes a meaningful additional benefit of the explanation panel.
Conversely, annotators perceived information-richer methods as more useful, revealing a performance--perception gap in hallucination-risk visualization.
An exploratory robustness probe with intentionally flipped hallucination labels provided preliminary evidence that unreliable visual cues may reduce quiz accuracy, although item and order effects cannot be separated from label flipping.
In this controlled task, the findings suggest that Information Richness should be calibrated for perceived usefulness as well as label reliability, cognitive cost, and accurate verification behavior.
Future work should test these patterns with more diverse annotators, documents, and detectors in more naturalistic RAG settings.

\section*{Limitations}

First, the experiments prioritized internal validity: they used fictional pseudo-documents, predefined queries, multiple-choice quiz questions, a five-minute limit, single-turn interactions, and at most one injected IH and one injected EH per sample.
Quiz accuracy therefore measures task-specific comprehension, not verification actions such as editing or rejecting a response, seeking external evidence, or making open-ended trust judgments.
The findings may differ with naturalistic corpora, retrieval failures, user-generated or open-ended queries, and multi-turn interaction.
Second, the 100-sample task set covers four injected hallucination configurations but a limited range of topics, formats, document lengths, and evidential ambiguity.
The UI was likewise limited to highlighting, IH/EH color coding, and short textual explanations; the C--D comparison evaluates the explanation panel as a whole and does not isolate the effects of its content, layout, quality, or reading demands.
Other interfaces, such as inline citations, source previews, or interactive evidence exploration, may produce different trade-offs.
Third, all cues came from a single detector, GPT-5, which was also used to generate the materials, and detector performance was evaluated against construction labels rather than independently adjudicated human labels.
Identical predictions across Methods B--D prevent detector differences from confounding the comparison, but the absolute usefulness of the cues depends on detector quality and calibration.
Verification behavior was measured mainly through reference-button clicks, so the behavioral interpretation in \S~\ref{subsec:gap} remains tentative.
Fourth, the performance--perception gap is a condition-level comparison between forced usefulness rankings and quiz-accuracy estimates; we did not directly test the within-annotator relationship between perceived usefulness and quiz accuracy, and the rankings do not capture absolute usefulness.
Fifth, Method E was a separately administered exploratory condition outside the Latin-square design of Methods A--D and was administered last for all annotators; because its comparisons did not use matched items, learning, fatigue, carry-over, and item effects cannot be separated from label flipping (\S~\ref{subsec:additional_experiment}).
Finally, we recruited 12 annotators in a within-subjects design.
Their 80 sample-level trials each support within-study comparisons but do not substitute for annotator diversity, particularly for conclusions about preference, trust, and behavioral mechanisms.
The main models included random intercepts but no random slopes for method, and the main accuracy model did not include a quiz-question random intercept; quiz-question-level variability was examined only in exploratory analyses.

\section*{Ethical Considerations}

Prior to conducting the experiments in this study, we obtained ethical approval from our organization (CAE-2025-08).
In addition, annotators in the experiment were either employees of our organization or were recruited through a crowdsourcing service.
To ensure fair compensation, we paid an hourly wage of JPY 2,000, which exceeds the minimum wage in Tokyo, Japan, as of October 3, 2025 (JPY 1,226).
Payment was calculated based on the maximum working time (8 hours).
If the maximum working time was reached, the experiment was terminated even if the annotator was still in the middle of the task.
Annotation was performed online on a web page. Detailed guidelines were made available for viewing on the web page.
This ensured transparency and ethical consistency.

% \section*{Acknowledgments}

% This research is the result of a joint study between CyberAgent, Inc., AI Shift, Inc., and Tohoku University.

% Bibliography entries for the entire Anthology, followed by custom entries
%\bibliography{custom,anthology-overleaf-1,anthology-overleaf-2}

% Custom bibliography entries only
\bibliography{j_yourrefs}

\appendix

\section{Distribution of Detector-Predicted Hallucination Patterns}
\label{sec:appendix_hallucination_distribution}

\begin{table*}[t]
    \centering
    \small
    \begin{tabular}{lrrrrrr}
        \hline
        Subset & IH:0, EH:0 & IH:1, EH:0 & IH:0, EH:1 & IH:1, EH:1 & Other & Total \\
        \hline
        $S_1$ & 3 & 4 & 5 & 4 & 4 & 20 \\
        $S_2$ & 8 & 4 & 3 & 3 & 2 & 20 \\
        $S_3$ & 2 & 5 & 7 & 4 & 2 & 20 \\
        $S_4$ & 2 & 4 & 4 & 7 & 3 & 20 \\
        $S_5$ & 4 & 5 & 7 & 3 & 1 & 20 \\
        \hline
    \end{tabular}
    \caption{
    Distribution of hallucination patterns across the 20-sample subsets $S_1$--$S_5$ based on the hallucination detector's sentence-level predictions used for visualization.
    ``Other'' groups samples with more complex prediction patterns (e.g., $\ge2$ IH and/or $\ge2$ EH sentences).
    }
    \label{tab:hallucination_distribution}
    \vspace{-2mm}
\end{table*}

Table~\ref{tab:hallucination_distribution} summarizes the distribution of detector-predicted hallucination patterns across the 20-sample subsets used in the experiment.
Although our data construction procedure intentionally inserted at most one intrinsic hallucination (IH) and at most one extrinsic hallucination (EH) into each response, the visualizations presented to annotators were generated from sentence-level predictions by the hallucination detection model.
As a result, some responses contained more complex predicted patterns than the injected hallucination structure.

In the table, ``IH:$k$, EH:$l$'' indicates that the response contains exactly $k$ sentences predicted as IH and exactly $l$ sentences predicted as EH, where $k,l\in\{0,1\}$.
The ``Other'' category groups responses with more complex prediction patterns, such as two or more IH and/or EH sentences.
Subsets $S_1$--$S_4$ were used in the main experiment, and $S_5$ was used in the exploratory robustness probe.

\begin{figure*}[b]
\centering
\includegraphics[width=\linewidth]{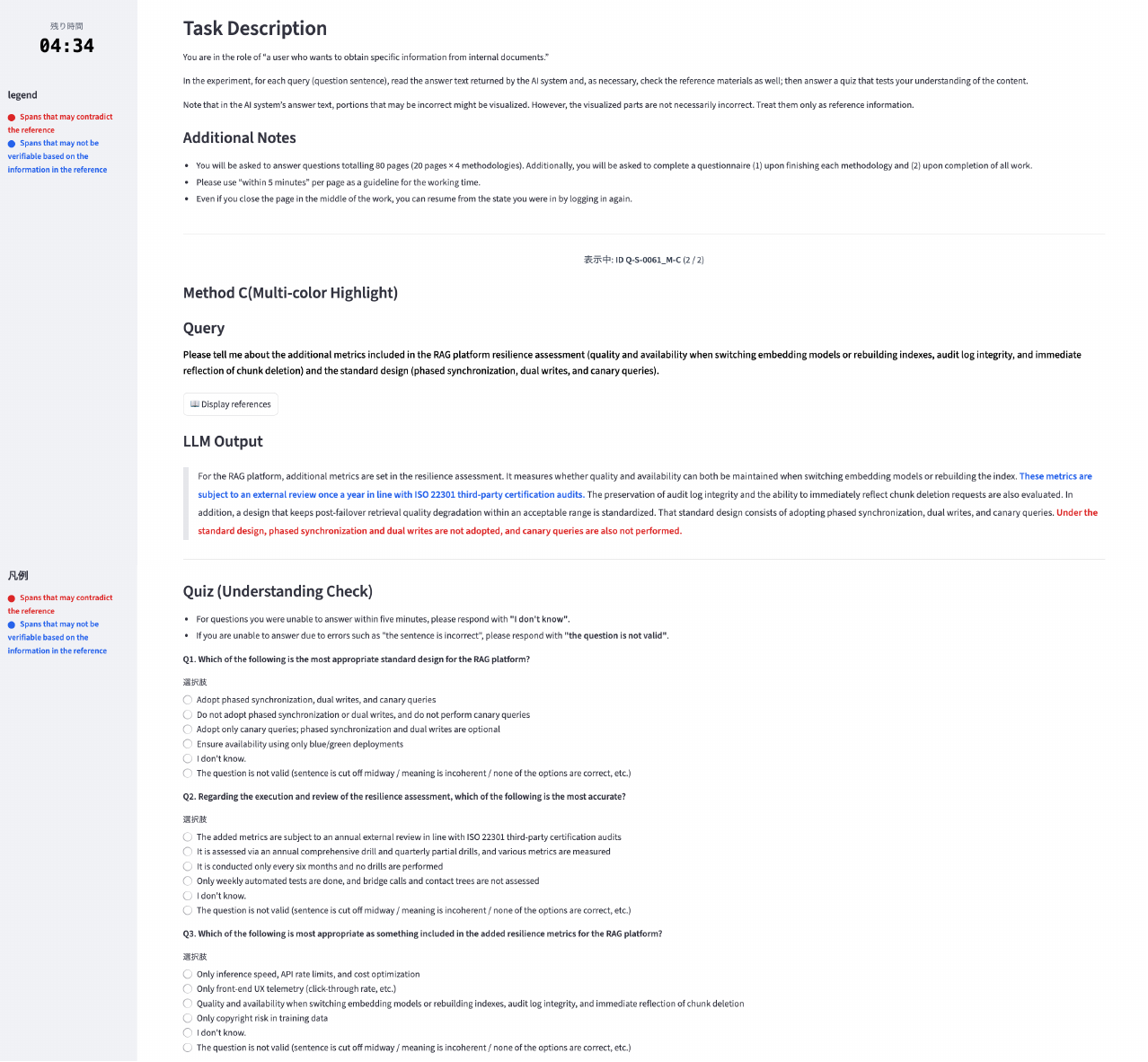}
\caption{Main annotation page for each sample.}
\label{fig:appendix_ui_main}
\end{figure*}

\section{Annotation Interface}
\label{sec:appendix_annotation_ui}

We show the annotation interface used in our user experiments (Figure~\ref{fig:appendix_ui_main}).
Figure~\ref{fig:appendix_ui_main} presents the annotation interface, including the main annotation page and the instruction.
The main page shows the remaining time, the query, a button to open the reference documents, and the LLM response with hallucination-risk visualization.
The instruction page summarizes the task explanation, workflow, and notes.
The legend indicates the meaning of each highlight color, and the interface also provides quiz instructions and multiple-choice questions per sample, including ``I don't know'' and ``Not answerable'' options.

\section{Visualization Examples}
\label{sec:appendix_visualization_examples}

\begin{figure*}[t]
\centering
\includegraphics[width=\linewidth]{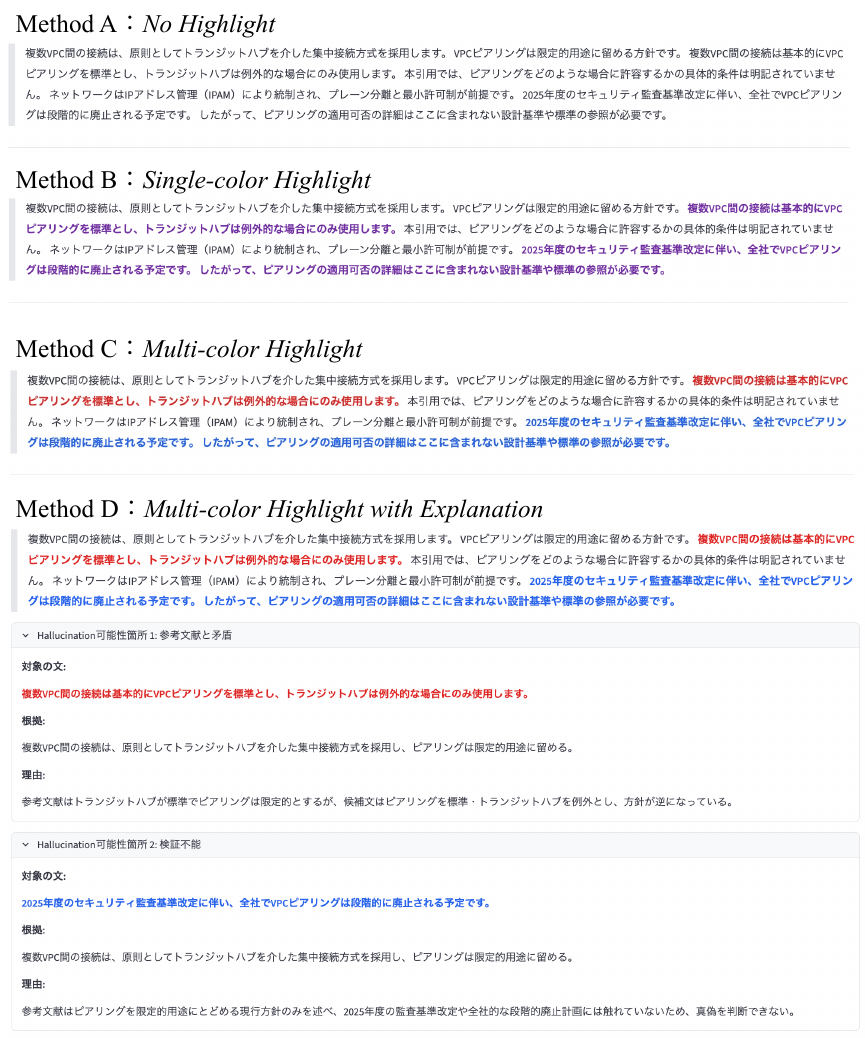}
\caption{
Japanese examples of the four visualization methods used in the experiment.
This version corresponds to the language of the actual user interface shown to annotators.
}
\label{fig:appendix_all_vis_jpn}
\end{figure*}

\begin{figure*}[t]
\centering
\includegraphics[width=\linewidth]{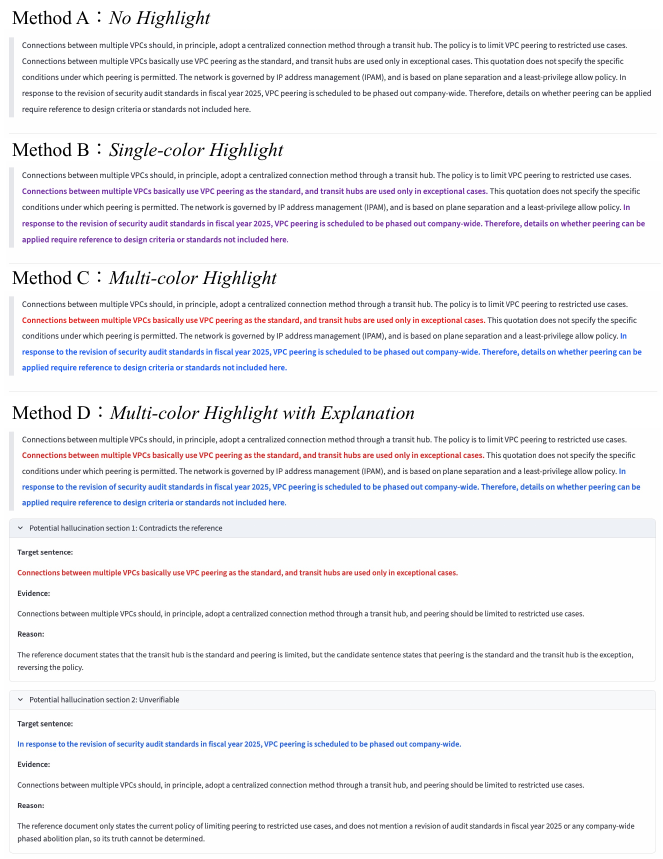}
\caption{
English examples of the four visualization methods.
This version is provided to help readers understand the visualization design.
}
\label{fig:appendix_all_vis_eng}
\end{figure*}

To provide a complete view of the visualization methods used in the experiment, we show examples of all visualization conditions in Japanese and English.
Figure~\ref{fig:appendix_all_vis_jpn} shows the Japanese version of the visualization examples, which corresponds to the interface actually used in the user experiment.
Figure~\ref{fig:appendix_all_vis_eng} shows the English version prepared for readers.

The four visualization methods are: A: \textit{No Highlight}, B: \textit{Single-color Highlight}, C: \textit{Multi-color Highlight}, and D: \textit{Multi-color Highlight with Explanation}.
A: \textit{No Highlight} presents the LLM response without explicit hallucination cues.
B: \textit{Single-color Highlight} adds localization by highlighting potentially hallucinated spans in a single color.
C: \textit{Multi-color Highlight} adds characterization by distinguishing intrinsic hallucinations (IH) from extrinsic hallucinations (EH) using different colors.
D: \textit{Multi-color Highlight with Explanation} further adds explanatory support by providing short textual explanations for the highlighted spans.

\section{Detailed Model Results for the Main Experiment}
\label{sec:appendix_model_details}

This appendix reports the trial-level mixed-model summaries that supplement the main-text claims in \S~\ref{subsec:RQ1} and \S~\ref{subsec:RQ2}.
In Table~\ref{tab:appendix_refview_model}, ``$\Delta$ vs.\ prev.''\ denotes the change in the descriptive mean relative to the immediately less rich method, ``IRR vs.\ A'' is the model-based effect with Method A as the reference, and ``Adj.''\ columns report adjacent contrasts (B/A, C/B, D/C) with Holm-corrected $p$-values for the three adjacent comparisons; Methods A--D are as defined in Figure~\ref{fig:methods_vis}.

Quiz accuracy was analyzed with a grouped-binomial GLMM with a logit link (lme4).
Each of the 960 annotator--sample observations (12 annotators $\times$ 80 samples) recorded the number of correct responses out of three quiz questions (2,880 responses in total):

{\small
\begin{verbatim}
Correct/3 ~ Method + Order
    + (1 | Annotator) + (1 | Sample)
\end{verbatim}
}
In all four main outcome models, Method was a categorical fixed effect and Order was the numeric presentation order (1--4); no random slopes or interactions were included.
Method used treatment coding with Method A as the reference, and Order was entered as an uncentered numeric covariate; adjacent contrasts were computed from estimated marginal means (emmeans).
The omnibus method effect was assessed by a likelihood-ratio test against a model omitting Method ($\chi^2(3){=}32.52$, $p<0.001$).
Descriptive means, OR estimates with 95\% CIs, and Holm-corrected adjacent contrasts are reported in Table~\ref{tab:RQ1_results}.

Reference-view counts per sample were analyzed with a negative binomial GLMM (NB2, log link; glmmTMB).
Because the sample random-intercept variance was estimated near zero and caused estimation problems (Appendix~\ref{sec:appendix_estimation_checks}), the final model retained only the annotator random intercept:

{\small
\begin{verbatim}
RefViews ~ Method + Order
    + (1 | Annotator)
\end{verbatim}
}
Table~\ref{tab:appendix_refview_model} reports means per 20 samples, whereas the model was fitted to counts per sample.
Evidence for method differences in reference views depended on the count-distribution assumption (Appendix~\ref{sec:appendix_estimation_checks}).
The omnibus method effect provided weak evidence ($\chi^2(3){=}7.32$, $p{=}0.06$), and the C--D adjacent contrast similarly provided weak evidence for fewer views under \textit{Multi-color Highlight with Explanation} than under \textit{Multi-color Highlight} (Holm-adjusted $p{=}0.06$), consistent with the descriptive pattern that source checking peaked under \textit{Multi-color Highlight} and dropped under \textit{Multi-color Highlight with Explanation}.

\begin{table}[t]
\centering
\small
\setlength{\tabcolsep}{4pt}
\resizebox{\columnwidth}{!}{%
\begin{tabular}{lcccc}
\hline
Method & Mean & $\Delta$ vs.\ prev.\ & IRR vs.\ A [95\% CI] & Adj.\ IRR ($p_{\text{Holm}}$) \\
\hline
A &  9.33 & --       & 1 (ref)           & -- \\
B & 10.83 & $+1.50$   & 1.15 [0.85, 1.57] & 1.15, $0.49$ \\
C & 12.25 & $+1.42$   & 1.33 [0.98, 1.80] & 1.15, $0.49$ \\
D &  9.08 & $-3.17$   & 0.99 [0.72, 1.37] & 0.75, $0.06$ \\
\hline
\end{tabular}
}
\caption{Reference-view counts per 20 samples: negative binomial GLMM results. 95\% CIs for IRRs vs.\ A are Bonferroni-adjusted for the three comparisons with Method A.}
\label{tab:appendix_refview_model}
\end{table}

Task time was analyzed in two parts (Table~\ref{tab:appendix_time_model_details}).
Timeout risk (an elapsed time of at least 300 seconds) was modeled for all 960 observations with a logistic GLMM, and completion time was modeled on the natural-log scale for the 855 non-timeout observations with a linear mixed model fitted by maximum likelihood (both lme4).
Both models used the same fixed effects and crossed random intercepts:

{\small
\begin{verbatim}
Timeout ~ Method + Order
    + (1 | Annotator) + (1 | Sample)
LogTime ~ Method + Order
    + (1 | Annotator) + (1 | Sample)
\end{verbatim}
}
Each omnibus test compared the fitted model with a model omitting Method.
Time ratios were obtained by exponentiating log-time contrasts and describe completion time conditional on not timing out.
The timeout-risk omnibus did not provide clear evidence for a method effect ($\chi^2(3){=}3.04$, $p{=}0.39$; observed rates A: 9.2\%, B: 9.2\%, C: 13.3\%, D: 12.1\%).
Among completed trials, the completion-time omnibus provided strong evidence for a method effect ($\chi^2(3){=}17.98$, $p<0.001$; mean times A: 152.8\,s, B: 153.2\,s, C: 161.5\,s, D: 165.7\,s), driven mainly by the B--C contrast; the A--B and C--D contrasts were not clear.
In Table~\ref{tab:appendix_time_model_details}, time ratios are reported as the first method relative to the second method (A/B, B/C, C/D), so values $<1$ indicate that the richer method took longer.

\begin{table}[t]
\centering
\small
\setlength{\tabcolsep}{6pt}
\begin{tabular}{lccc}
\hline
Adjacent contrast & Time ratio & 95\% CI & $p_{\text{Holm}}$ \\
\hline
A/B & 1.02 & [0.95, 1.10] & $0.998$ \\
B/C & 0.90 & [0.84, 0.98] & $0.005$ \\
C/D & 0.99 & [0.92, 1.07] & $0.998$ \\
\hline
\end{tabular}
\caption{Adjacent time ratios from the log-time linear mixed model on completed trials. 95\% CIs are Bonferroni-adjusted for the three adjacent contrasts.}
\label{tab:appendix_time_model_details}
\end{table}

\subsection{Exploratory quiz-question analyses}
\label{sec:appendix_exploratory_models}
The allocation analysis modeled 2,880 individual responses to 240 quiz questions with a binomial GLMM (logit link), with Method and Block as categorical fixed effects. The Holm-adjusted C--D comparison gave $p=0.803$.

A separate binomial GLMM modeled errors under Method D as a function of difficulty estimated from Methods A--C, defined as one minus accuracy under A--C in units of 0.1. Higher estimated difficulty was associated with more errors under D ($p<0.001$). These exploratory models differ from the main grouped-binomial analysis and do not propagate uncertainty in the estimated difficulty covariate.

{\small
\begin{verbatim}
Correct ~ Method + Block
    + (1 | Annotator) + (1 | QuizQuestion)
ErrorD ~ Difficulty
    + (1 | Annotator) + (1 | QuizQuestion)
\end{verbatim}
}

\subsection{Exploratory Method E models}
\label{sec:appendix_method_e_models}
Within Method E, each outcome was modeled with the same family and random-effect structure as the corresponding main model, replacing Method with label status (reference: non-flipped). Label status was tested with a likelihood-ratio test.

{\small
\begin{verbatim}
Correct/3 ~ LabelStatus + Order
    + (1 | Annotator) + (1 | Sample)
RefViews ~ LabelStatus + Order
    + (1 | Annotator)
Timeout ~ LabelStatus + Order
    + (1 | Annotator) + (1 | Sample)
LogTime ~ LabelStatus + Order
    + (1 | Annotator) + (1 | Sample)
\end{verbatim}
}

\subsection{Usefulness rankings}
\label{sec:appendix_rank_models}
Usefulness rankings were analyzed with a Friedman test blocking by annotator. A one-sided exact Page test assessed the ordered preference $A < B < C < D$ on scores transformed as $5-\text{rank}$. The Friedman test yielded $\chi^2$(3) = 18.70 ($p<0.001$), and the Page statistic was L = 340 ($p<0.001$).

\subsection{Estimation and diagnostics}
\label{sec:appendix_estimation_checks}
Analyses were run in R 4.5.2 with lme4 2.0-1, glmmTMB 1.1.14, ordinal 2025.12-29, and emmeans 2.0.2; completion-time contrasts used Kenward--Roger degrees of freedom (pbkrtest 0.5.5). lme4 models used bobyqa followed by Nelder--Mead (GLMMs) or nloptwrap (LMMs); glmmTMB and ordinal models used nlminb, the latter with 10 adaptive quadrature points. Apart from the reference-count models described below, all full and reduced models converged with positive-definite Hessians and non-singular random-effect estimates. For reference counts, the initial models with annotator and sample random intercepts had near-zero sample variance and non-positive-definite Hessians, so the sample intercept was removed. The Method E count model still issued a false-convergence warning, but refits with another optimizer and with a Poisson distribution left the fixed-effect estimates essentially unchanged.

Residual checks suggested underdispersion in reference counts and unequal residual variance in log times. We therefore ran post hoc sensitivity analyses with Conway--Maxwell--Poisson count models and log-time models whose residual variance depended on method and order, keeping the original mean and random-effect structures (Table~\ref{tab:diagnostic_sensitivity}). Evidence for A--D reference-count differences depended on the distributional assumption, whereas A--D completion-time evidence remained strong. Full diagnostic output is provided with our analysis code.

\begin{table*}[t]
\centering
\small
\begin{tabular}{lrlllr}
\hline
Outcome & $N$ & Original model & $p$ & Sensitivity model & $p$ \\
\hline
A--D reference counts & 960 & NB2 & 0.062 & COM-Poisson & 0.013 \\
E reference counts & 240 & NB2 & 0.542 & COM-Poisson & 0.448 \\
A--D log time & 855 & Constant variance & $<0.001$ & Varying variance & $<0.001$ \\
E log time & 212 & Constant variance & 0.198 & Varying variance & 0.197 \\
\hline
\end{tabular}
\caption{Post hoc distributional sensitivity checks. Likelihood-ratio tests have df=3 for A--D and df=1 for E. Varying-variance models use condition and numeric order in the dispersion submodel; full and reduced fits retain the same dispersion structure. Original count models use NB2 with an annotator random intercept. COM-Poisson denotes the Conway--Maxwell--Poisson distribution.}
\label{tab:diagnostic_sensitivity}
\end{table*}

\section{Analysis by Hallucination Type}
\label{sec:appendix_type_analysis}

To further examine how the visualization methods affected different kinds of quiz questions, we conducted a descriptive analysis by hallucination type.
Each quiz question was categorized according to whether it referred to a statement involving an intrinsic hallucination (IH), an extrinsic hallucination (EH), or no hallucinated content.
Specifically, we grouped quiz questions into: (i) \textit{IH-related} questions referring to intrinsically hallucinated content; (ii) \textit{EH-related} questions referring to extrinsically hallucinated content; and (iii) \textit{None} questions referring to non-hallucinated content involving neither hallucination type.
For each annotator and each visualization method, we first computed the mean accuracy within each question group, and then averaged these values across annotators.
The \emph{All} column reports the overall accuracy across all quiz questions.

\begin{table}[t]
\centering
\small
\begin{tabular}{lcccc}
\hline
Method & \textit{IH-related} & \textit{EH-related} & \textit{None} & \textit{All} \\
\hline
A& 0.701 & 0.721 & 0.899 & 0.781 \\
B& 0.800 & 0.800 & 0.912 & 0.844 \\
C& 0.850 & 0.812 & 0.929 & 0.875 \\
D& 0.826 & 0.824 & 0.933 & 0.869 \\
\hline
\end{tabular}
\caption{
Quiz accuracy by visualization method and hallucination type (Methods A--D as defined in Figure~\ref{fig:methods_vis}).
Values were computed by first averaging accuracy within each annotator and method, and then averaging across annotators.
The \textit{None} column includes only quiz questions that did not refer to contradictory or unverifiable hallucinated content.
}
\label{tab:appendix_type_accuracy}
\end{table}

As shown in Table~\ref{tab:appendix_type_accuracy}, the overall accuracy showed the same pattern as the main analysis: accuracy improved from A: \textit{No Highlight} to B: \textit{Single-color Highlight} and C: \textit{Multi-color Highlight}, while the estimated benefit did not increase further at D: \textit{Multi-color Highlight with Explanation}.
The type-specific results suggest that this improvement was especially pronounced for hallucination-related questions.
For \textit{IH-related} questions, accuracy increased substantially from A to C ($0.701 \rightarrow 0.850$), suggesting that adding localization and characterization helped users accurately identify information that contradicted the source.
For \textit{EH-related} questions, accuracy also improved from A to D ($0.721 \rightarrow 0.824$), although the improvement was more gradual than for \textit{IH-related} questions.
In contrast, \textit{None} questions showed high accuracy across all methods ($0.899$--$0.933$), suggesting that these questions were comparatively easier and less dependent on hallucination visualization.

Overall, this analysis suggests that increasing Information Richness mainly improved accuracy for hallucination-related questions, especially for \textit{IH-related} content.
However, D: \textit{Multi-color Highlight with Explanation} did not clearly outperform C: \textit{Multi-color Highlight}, consistent with the main finding that localization and characterization provided a comparatively well-balanced design, whereas adding explanatory support did not necessarily yield further gains in information acquisition.

\begin{figure}[t]
\centering
\includegraphics[width=\linewidth]{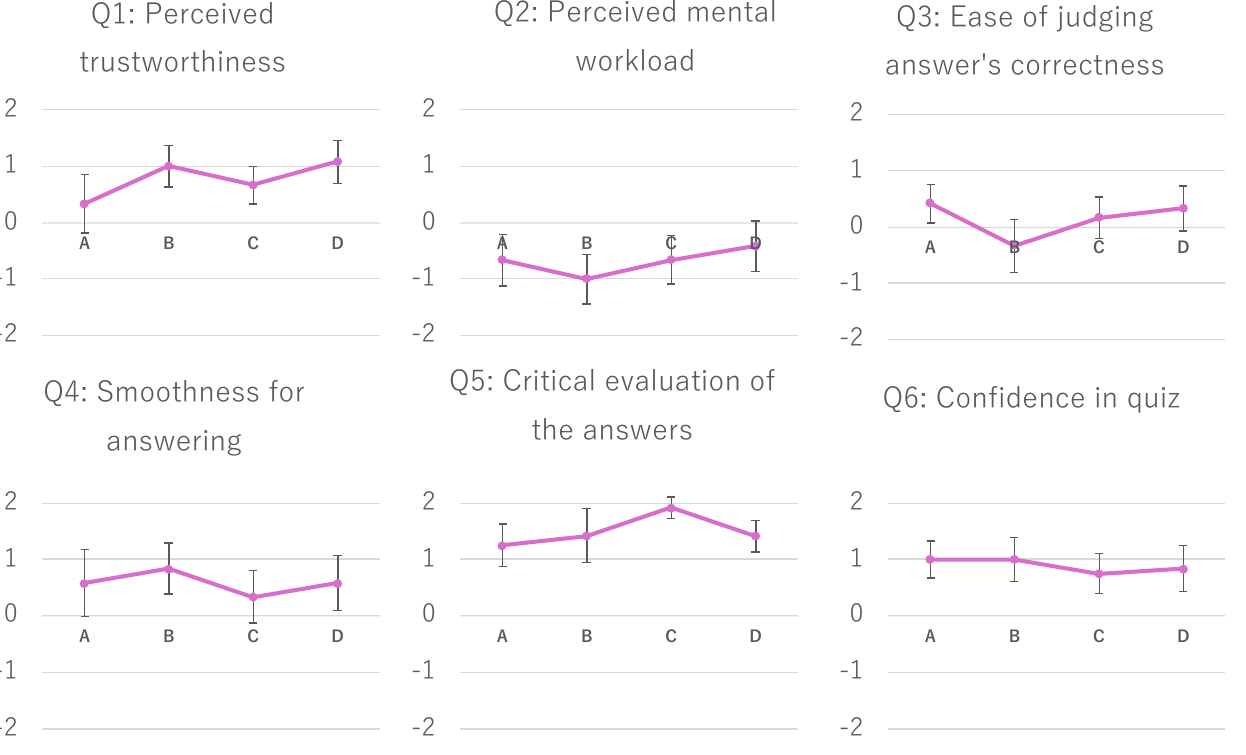}
\caption{
Results of the during-experiment questionnaire in the main experiment.
Scores are mean coded Likert ratings expressed as deviations from the neutral midpoint ($4$) on the 1--7 scale.
For Q1, Q3, Q4, Q5, and Q6, positive values indicate more favorable evaluations; for Q2, positive values indicate higher perceived workload.
}
\label{fig:questionnaire}
\end{figure}

\begin{figure}[t]
\centering
\includegraphics[width=\linewidth]{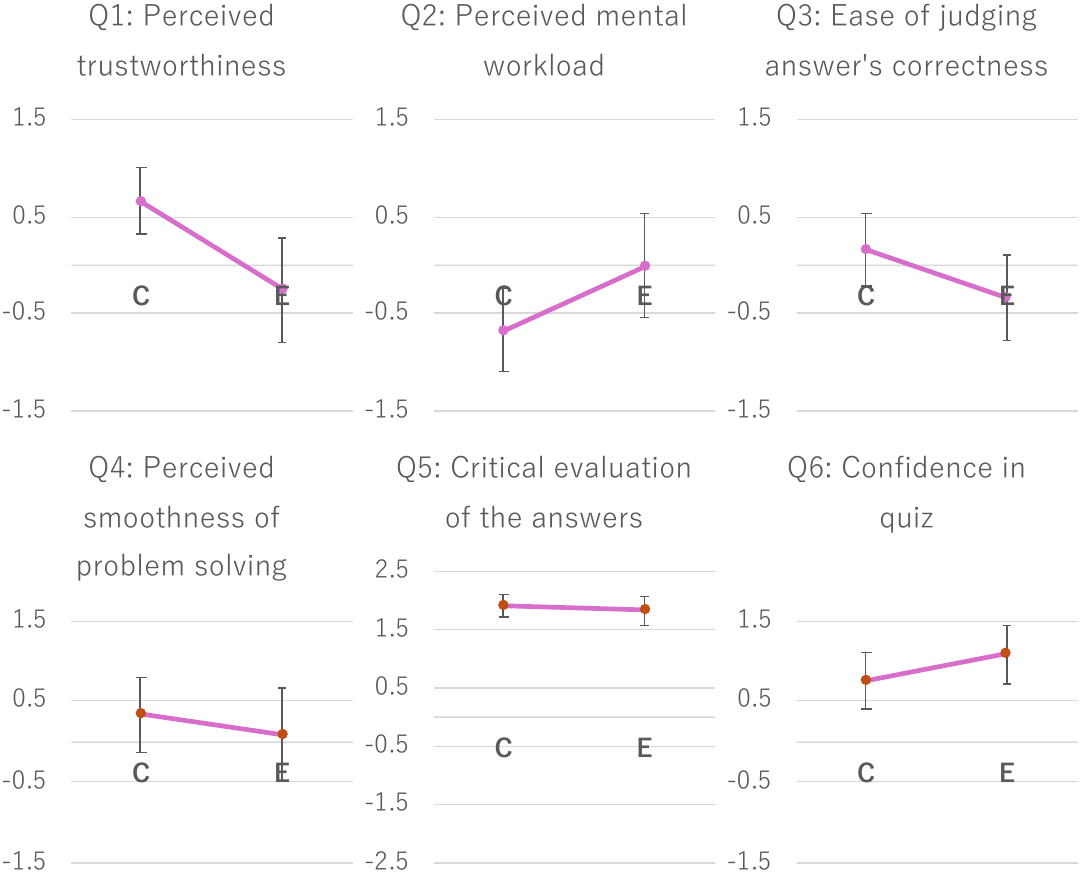}
\caption{
During-experiment questionnaire results for the exploratory robustness probe (Method C as defined in Figure~\ref{fig:methods_vis}; Method E: \textit{Multi-color Highlight with Flipped Labels}).
}
\label{fig:ex2_questionnaire}
\end{figure}

\section{During-Experiment Questionnaire Results}
\label{sec:appendix_questionnaire_results}

Figure~\ref{fig:questionnaire} shows the during-experiment questionnaire results for the main experiment, complementing the coded-scale means and SDs reported in Table~\ref{tab:RQ2_questionnaire}.
The during-experiment questionnaire was treated as supplementary evidence for RQ1 and RQ2.
Method effects on each of Q1--Q6 were assessed with cumulative-logit mixed models (ordinal), using 48 annotator--method observations from 12 annotators:

{\small
\begin{verbatim}
Score ~ Method + Order
    + (1 | Annotator)
\end{verbatim}
}
Here, Order was categorical.
Each omnibus test compared the full model with a model omitting Method; these tests were not adjusted across questionnaire questions.
For analysis, the original 1--7 responses were reversed (8 minus the response).
For the descriptive presentation in Figure~\ref{fig:questionnaire}, Table~\ref{tab:RQ2_questionnaire}, and the within-method means quoted in the main text, scores were coded so that higher values indicate more favorable evaluations for Q1, Q3, Q4, Q5, and Q6, and higher perceived workload for Q2.
Table~\ref{tab:RQ2_questionnaire} reports these coded scores on the 1--7 scale, whereas Figure~\ref{fig:questionnaire} expresses the same means as deviations from the neutral midpoint ($4$).
The main-text questionnaire interpretation focuses on Q1, Q2, Q3, and Q5; Q4 and Q6 were treated as exploratory.
Q7 was a categorical question about reference-text reliance and is reported descriptively only.

Figure~\ref{fig:ex2_questionnaire} shows the during-experiment questionnaire results for the exploratory robustness probe comparing Method C: \textit{Multi-color Highlight} and Method E: \textit{Multi-color Highlight with Flipped Labels}.
These results supplement the analysis of how incorrect hallucination labels affected subjective evaluations.

\section{Items in the During-Experiment and Post-Experiment Questionnaires}
\label{sec:questionnaire_question}

\paragraph{During-experiment questionnaire}
The during-experiment questionnaire was designed by drawing on items from the NASA Task Load Index (NASA-TLX), a measure of subjective workload~\cite{hancock_development_1988}, and from questions used in prior studies on hallucination-visualization user interfaces by~\citet{do2024facilitating_chiws} and~\citet{hill}.
Items Q1--Q6 used a 7-point Likert scale ranging from ``Strongly agree'' to ``Strongly disagree.''
Item Q7 used a 7-point ordered categorical scale on reference reliance, anchored from response-text-only to references-only.

\begin{itemize}
\item Q1: I felt that I could trust this AI system in solving the problems.
    \begin{itemize}
        \item Strongly agree
        \item Somewhat agree
        \item Slightly agree
        \item Neither agree nor disagree
        \item Slightly disagree
        \item Somewhat disagree
        \item Strongly disagree
    \end{itemize}
\item Q2: Using this AI system, I felt that the concentration and mental load required to work on the task were low.
    \begin{itemize}
        \item (Same as Q1)
    \end{itemize}
\item Q3: It was easy to judge whether the answer text produced by this AI system was correct.
    \begin{itemize}
        \item (Same as Q1)
    \end{itemize}
\item Q4: Using this AI system, I felt that I could proceed smoothly in solving the problems.
    \begin{itemize}
        \item (Same as Q1)
    \end{itemize}
\item Q5: Rather than accepting the AI system's answers at face value, I was able to view them critically, such as by checking the references when necessary.
    \begin{itemize}
        \item (Same as Q1)
    \end{itemize}
\item Q6: I was able to answer the problems with confidence.
    \begin{itemize}
        \item (Same as Q1)
    \end{itemize}
\item Q7: When solving the quiz (comprehension check), which did you refer to more: the answer text or the references?
    \begin{itemize}
        \item I based my answer almost entirely on the response text and hardly looked at the references.
        \item I mainly relied on the response text and did not look at the references much.
        \item I tended to rely more on the response text.
        \item I relied on the response text and the references to about the same extent.
        \item I tended to rely more on the references.
        \item I mainly relied on the references and did not look at the response text much.
        \item I based my answer almost entirely on the references and hardly looked at the response text.
    \end{itemize}
\end{itemize}

\paragraph{Post-experiment questionnaire}
For the post-experiment questionnaire, we prepared questions regarding how useful each visualization method was, the difficulty of the problems, and annotators' use of generative AI.
Item Q1 was a forced ranking of the four visualization methods by usefulness.
Items Q3 and Q4 used a 7-point Likert scale (``Strongly agree''--``Strongly disagree''), and items Q5 (verification frequency) and Q6 (LLM usage frequency) used a 5-point frequency scale.
Items Q2 and Q7 were open-ended.

\begin{itemize}
\item Q1: Please rank the AI systems in order of how useful they were.
    \begin{itemize}
        \item (Annotators will rank each visualization method from 1st to 4th place in order of usefulness.)
    \end{itemize}
\item Q2: Please explain the reason for the above ranking.
    \begin{itemize}
        \item (Annotators will provide a free-form description of the reason for Q1.)
    \end{itemize}
\item Q3: The problems were difficult.
    \begin{itemize}
        \item Strongly agree
        \item Somewhat agree
        \item Slightly agree
        \item Neither agree nor disagree
        \item Slightly disagree
        \item Somewhat disagree
        \item Strongly disagree
    \end{itemize}
\item Q4: All problems were about the same level of difficulty.
    \begin{itemize}
        \item (Same as Q3)
    \end{itemize}
\item Q5: When you use conversational AI such as ChatGPT, to what extent do you verify (fact-check) the information in its responses?
    \begin{itemize}
        \item Every time
        \item Quite often
        \item Sometimes
        \item Rarely
        \item Never
    \end{itemize}
\item Q6: Over the past three months, how much have you used conversational AI (LLMs) such as ChatGPT?
    \begin{itemize}
        \item Almost every day
        \item A few times a week
        \item About once a week
        \item About once a month
        \item Less than once a month
    \end{itemize}
\item Q7: If you have any other comments or questions, please feel free to write them.
    \begin{itemize}
        \item (Annotators may freely write comments or questions in the open-ended format.)
    \end{itemize}
\end{itemize}

\section{Main Topics of the Pseudo-Document Corpus}
\label{sec:appendix_corpus_contents}

The following list presents the 20 main topics covered by the fictional internal-company document corpus described in Section~\ref{subsec:data_preparation}.

\begin{enumerate}[label=\arabic*., leftmargin=*, nosep]
    \item Company Overview and Organization
    \item Information Security and Compliance
    \item IT Infrastructure and Cloud Platforms
    \item Development Processes and DevOps
    \item Product Design and Architecture
    \item Generative AI and RAG Platform
    \item Operations and SRE
    \item Sales and Customer Success
    \item Support and Knowledge Base
    \item Human Resources, General Affairs, and Legal
    \item Marketing and Public Relations
    \item Quality Assurance (QA) and Test Strategy
    \item Data Governance and Privacy
    \item Risk Management and Internal Audit
    \item Procurement and Vendor Management
    \item Financial Management and Revenue Recognition
    \item Business Continuity Planning (BCP) and Crisis Management
    \item International Expansion and Regional Operations
    \item Research and Development (R\&D) and Technology Strategy
    \item AI Ethics and Responsible AI Use
\end{enumerate}

\end{document}